\documentclass[letterpaper, 10 pt, conference]{ieeeconf}  

\IEEEoverridecommandlockouts                              

\usepackage{graphicx, siunitx, subcaption}
\usepackage{fontawesome}
\usepackage{graphicx}
\usepackage{booktabs}
\usepackage{amsmath, stackengine, dsfont, amssymb, pifont, graphicx}
\usepackage{booktabs}
\usepackage{multirow, bbold, soul}
\usepackage{booktabs}
\usepackage{amsmath, stackengine, dsfont, amssymb, pifont, graphicx, amsfonts}
\usepackage{booktabs}
\usepackage{multirow, bbold, soul}

\usepackage[accsupp]{axessibility}  

\usepackage{xcolor,colortbl}
\usepackage{multirow}
\usepackage{amsmath}

\usepackage{color}
\definecolor{crimson}{rgb}{0.86, 0.08, 0.24}
\definecolor{gray}{rgb}{0.5,0.5,0.5}
\definecolor{green}{rgb}{0, 0.4, 0}
\definecolor{mahogany}{rgb}{0.75, 0.25, 0.0}
\definecolor{purple}{rgb}{0.6, 0, 0.6}
\definecolor{darkgreen}{rgb}{0, 0.4, 0}
\definecolor{frenchblue}{rgb}{0.0, 0.45, 0.73}

\title{\LARGE \bf
Enhancing Single Image to 3D Generation using Gaussian Splatting and Hybrid Diffusion Priors
}

\author{Hritam Basak$^{1,3,\dagger}$, Hadi Tabatabaee$^{1,\dagger}$, Shreekant Gayaka$^1$, Ming-Feng Li$^{1,2}$, Xin Yang$^1$, \\Cheng-Hao Kuo$^1$, Arnie Sen$^1$,  Min Sun$^1$, Zhaozheng Yin$^{3}$\\
{$^1$Amazon Lab126, $^2$Carnegie Mellon University, $^3$Stony Brook University,  }
\thanks{$\dagger$ Equal contribution}
}

\begin{document}

\maketitle
\thispagestyle{empty}
\pagestyle{empty}

\begin{abstract}
3D object generation from a single image involves estimating the full 3D geometry and texture of unseen views from an unposed RGB image captured in the wild. Accurately reconstructing an object's complete 3D structure and texture has numerous applications in real-world scenarios, including robotic manipulation, grasping, 3D scene understanding, and AR/VR.
Recent advancements in 3D object generation have introduced techniques that reconstruct an object's 3D shape and texture by optimizing the efficient representation of Gaussian Splatting, guided by pre-trained 2D or 3D diffusion models. However, a notable disparity exists between the training datasets of these models, leading to distinct differences in their outputs. While 2D models generate highly detailed visuals, they lack cross-view consistency in geometry and texture. In contrast, 3D models ensure consistency across different views but often result in overly smooth textures.
To address this limitation, we propose bridging the gap between 2D and 3D diffusion models by integrating a two-stage frequency-based distillation loss with Gaussian Splatting. Specifically, we leverage geometric priors in the low frequency spectrum from a 3D diffusion model to maintain consistent geometry, and use 2D diffusion model to refine the fidelity and texture in the high frequency spectrum of the generated 3D structure, resulting in more detailed and fine-grained outcomes. Our approach enhances geometric consistency and visual quality, outperforming the current SOTA. Additionally, we demonstrate the easy adaptability of our method for efficient object pose estimation and tracking.

\end{abstract}

\section{INTRODUCTION}\label{introduction}

\begin{figure}[t!]
    \centering
    \includegraphics[width=\columnwidth]{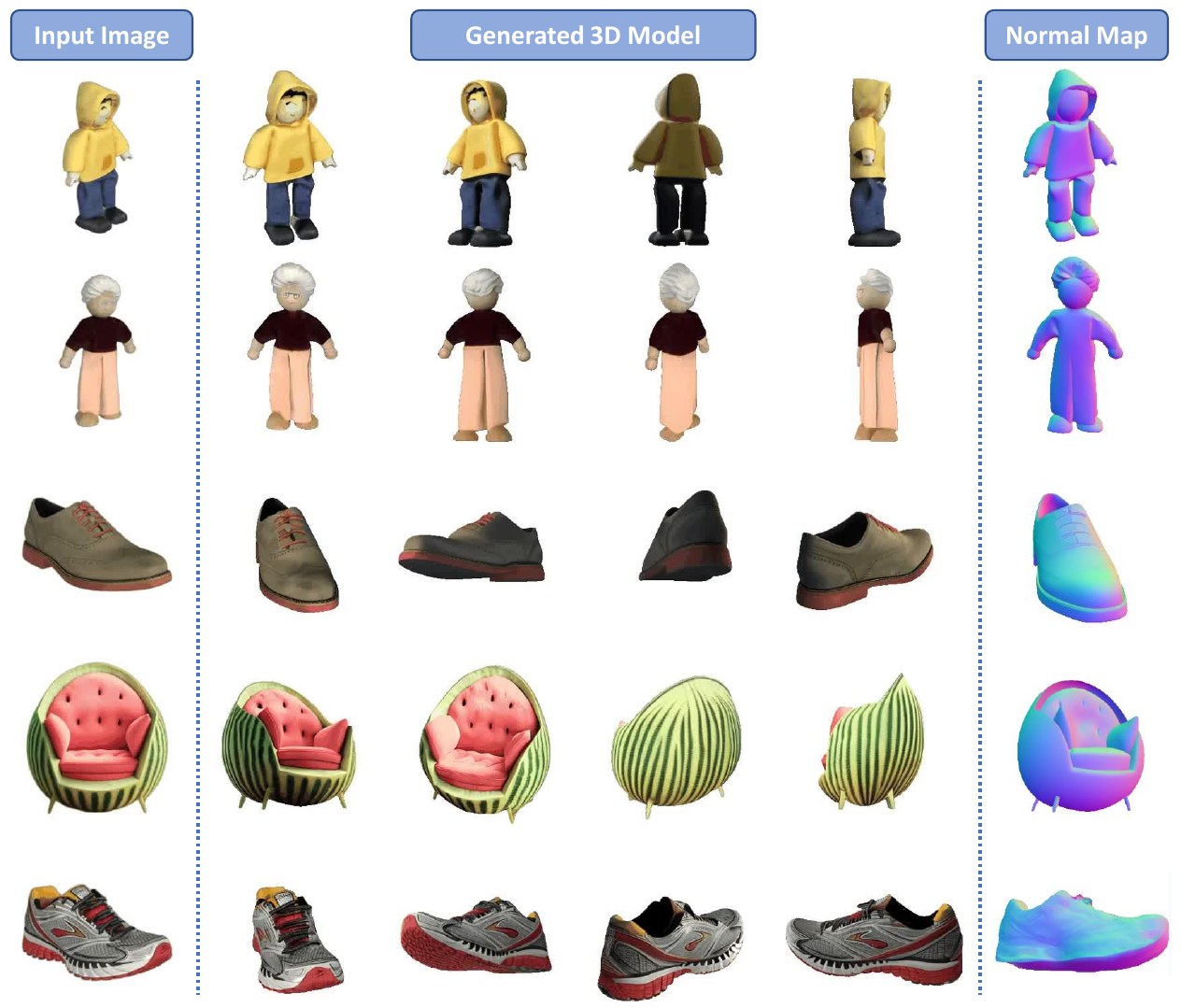}
    \caption{We improve single image to 3D generation, maintaining both geometric consistency and superior texture: the input image (col. 1), rendered multi-view images (col. 2-5), generated normal map (col. 6). We encourage readers check \textcolor{red}{supplementary file} for \textcolor{red}{video results}.}
    \label{fig:our_result}
    \vspace{-4.4mm}
\end{figure}

Recent advancements in 3D object generation leverage large pre-trained 2D and 3D diffusion models to reconstruct an object’s 3D shape and texture from a single image \cite{long2024wonder3d,tang2023dreamgaussian}. This involves initializing a 3D representation, such as Neural Radiance Fields (NeRF)\cite{mildenhall2021nerf} or 3D Gaussian Splatting (3DGS)\cite{kerbl20233dgs}, with a single 2D image and then optimizing it using 2D priors from diffusion models trained on large-scale datasets like LAION~\cite{schuhmann2021laion}. For example, DreamFusion~\cite{poole2022dreamfusion} uses Score Distillation Sampling (SDS) to distill 2D priors from pre-trained 2D diffusion models to optimize NeRF representations for 3D object generation. Other methods generate novel view images from a single 2D image using 2D diffusion models, which are then used for 3D reconstruction \cite{shi2023mvdream}. Although NeRF-based methods often require significant optimization time, recent approaches incorporating 3DGS~\cite{yi2023gaussiandreamer} have reduced this duration, though several minutes of optimization are still needed. Additionally, single-view 2D diffusion models lack inherent multi-view consistency, leading recent works~\cite{woo2024harmonyview} to stress the importance of maintaining this consistency for coherent 3D reconstructions. 
This advancement is critically important in robotics, where accurate 3D object reconstruction has numerous applications, and the community has recently intensified its focus on integrating these techniques to enhance robotic perception and manipulation \cite{kasahara2024ric,pan2024many,smitt2023pag}.


Two major unsolved challenges of 3D generation are \textit{visual fidelity} (sharp and crisp outputs, well-defined textures) and \textit{geometric realism} (geometrically consistent objects, that match our real-life perception). 
Some recent efforts have fine-tuned existing 2D image diffusion models on 3D datasets, thereby enabling 3D geometric understanding \cite{liu2023zero,shi2023mvdream}. Foundation models are another promising direction \cite{hong2023lrm,tang2024lgm}, enabling realistic 3D generation, thanks to large-scale paired datasets \cite{deitke2023objaverse} used for 2D-to-3D lifting. However, despite the ability of 3D-native methods to generate 3D-consistent assets—such as meshes, Gaussians, point clouds, or NeRF, sometimes within seconds, these methods continue to grapple with limitations in visual fidelity. Other works \cite{qian2023magic123} roughly combine 3D and 2D representations in the spatial domain, blatantly ignoring their distinctive outputs and geometric differences. This leads to a lack of realism (e.g., assymetric structure, missing parts, etc.), and blurry reconstructions \cite{li2024advances}.

Fourier123 \cite{yang2024fourier123}, on the other hand, observed a discrepancy in the outputs of 2D and 3D prior-based models (Stable Diffusion \cite{rombach2022high} and Zero123 \cite{liu2023zero}, respectively), both in image and frequency domains, and proposed a solution by proposing a 2D frequency-based SDS loss. However, they still fail to effectively combine the structural attributes from 2D priors and geometrical information from 3D prior, as visible in their findings. We draw motivation from this observation and propose a superior frequency-based combination of 2D and 3D priors. As highlighted in \cite{wu2023freeinit}, geometric features can be attributed to low-frequency components whereas high-frequency components are majorly responsible for textures and edges. We, therefore propose two novel SDS loss functions utilizing 3D and 2D priors (details in \ref{hf-sds-section}). Instead of utilizing the entire frequency distribution (as in \cite{yang2024fourier123}), we filter the low and high-frequency components to selectively prioritize geometry and texture from 3D and 2D priors.

Unlike \cite{yang2024fourier123}, we propose combining 2D and 3D guidance in the frequency domain for two novel distillation losses, enabling both the textural and geometric guidance in our pipeline. Specifically, our contributions are as follows:
\begin{itemize}
    \item We present a novel two-stage image-to-3D generation pipeline that combines 2D and 3D priors to achieve high-quality, geometrically consistent 3D models in under a minute, significantly advancing both fidelity, geometric realism, and speed for asset generation.
    \item We develop two novel frequency-based SDS losses: rather than distillation losses in the image domain, we utilize the low and high frequency components in the Fourier space of the 3D and 2D priors, respectively, and combine them to propose a novel hybrid frequency score distillation loss (hf-SDL). 
    \item Upon evaluation, our work outperforms existing SOTA models for single image to 3D generation tasks on three publicly available datasets, both in terms of 2D texture and 3D geometric metrics. 
    \item To further explore its robustness, we validate the generated 3D assets in pose estimation and tracking, useful for robotic applications in limited view scenarios. 
\end{itemize}

\section{RELATED WORK}\label{related_works}

\subsection{2D Diffusion Models for 3D Asset Generation}
Recent strides in 2D diffusion models \cite{ho2020denoising,croitoru2023diffusion} and expansive vision-language frameworks like CLIP \cite{radford2021learning} have revolutionized the generation of 3D assets by capitalizing on the sophisticated priors embedded in these 2D models. Seminal works such as DreamFusion \cite{poole2022dreamfusion} and SJC \cite{wang2023score} have showcased the capability of extracting 3D shapes from textual or visual cues through the distillation of 2D text-to-image generation models. These approaches generally involve optimizing 3D representations—such as Neural Radiance Fields (NeRF), meshes, or Signed Distance Functions (SDF)—and employing neural rendering to generate 2D images from multiple viewpoints. These images are then refined using 2D diffusion models or CLIP, guiding the optimization of 3D shapes through Score Distillation Sampling (SDS) \cite{melas2023realfusion,raj2023dreambooth3d,lin2023magic3d,tang2023make}. Nevertheless, these methods often encounter inefficiencies, including extended per-shape optimization times and challenges like the multi-face problem, largely due to the absence of explicit 3D supervision. Despite recent advances like One-2-3-45 \cite{liu2023one2345}, which attempt to enhance the process through generalizable neural reconstruction techniques, the results frequently fall short in geometric fidelity and detail.

\subsection{Multi-view Diffusion for Mesh Reconstruction}
Mesh reconstruction has remained a central task in graphics and 3D computer vision, given its wide usage in popular 3D engines like Blender and Maya. To achieve consistent multi-view image generation and mesh reconstruction, several approaches \cite{gu2023nerfdiff,deng2023nerdi,tang2024mvdiffusion++,tseng2022cla} have extended 2D diffusion models from single-view outputs to multi-view settings. However, these methods primarily target image generation and are not optimized for 3D reconstruction. Some techniques \cite{zhang2024text2nerf,xiang20233d} attempt to warp depth maps to generate incomplete novel views, followed by inpainting, but their effectiveness is hampered by inaccuracies in the initial depth maps generated by external models. Recent works such as Viewset Diffusion \cite{szymanowicz2023viewset}, SyncDreamer \cite{liu2023syncdreamer}, and MVDream \cite{shi2023mvdream} have employed attention layers to enhance multi-view consistency in color images. However, unlike normal maps that inherently encode geometric information, color-based reconstructions suffer from texture ambiguity, leading to difficulties in capturing fine geometric details or imposing significant computational demands. For instance, SyncDreamer necessitates dense views for 3D reconstruction, yet it still produces low-quality geometry with blurred textures. Similarly, MVDream relies on a time-intensive optimization process using Score Distillation Sampling (SDS), with its multi-view distillation process taking up to 1.5 hours. In contrast, our approach achieves high-quality textured mesh reconstruction in $<1$ minute, significantly reducing the computational burden while maintaining detailed geometry.

\subsection{3D Generative Models}
In contrast to the per-shape optimization guided by 2D diffusion models, several works have focused on directly training 3D diffusion models based on various 3D representations, including point clouds \cite{vahdat2022lion,zhou20213d}, meshes \cite{gao2022get3d,liu2023meshdiffusion}, and neural fields \cite{ntavelis2023autodecoding,muller2023diffrf,erkocc2023hyperdiffusion,gu2024control3diff,chen2023single,anciukevivcius2023renderdiffusion}. Despite these efforts, the limited availability of large-scale 3D datasets has restricted the validation of these models to a narrow range of shapes, posing challenges in scaling them up for broader applications. 
With very recent development of large-scale 3D datasets like Objaverse-XL \cite{deitke2023objaverse}, efforts have been made to train large reconstruction models for single image to 3D generation using transformer \cite{hong2023lrm}, triplane representation \cite{zou2024triplane}, flexicubes \cite{xu2024instantmesh}, Gaussian splatting \cite{tang2024lgm}, and many more. However, training a foundation model is compute-heavy and time-consuming, and hence, inaccessible for all. Moreover, due to smaller size of 3D datasets, as compared to publicly available 2D images \cite{schuhmann2022laion}, effective utilization of pretrained models on 2D datasets still remain an open problem. Additionally, to address the challenge of generating multi-view consistent images, recent methods have extended 2D diffusion models to produce consistent multi-view outputs \cite{liu2023zero,shi2023zero123++}. However, these methods struggle with low-quality geometry and blurring textures due to the inherent challenges of reconstructing 3D structures.

\begin{figure*}[h]
    \centering
    \includegraphics[height=70mm,width=1.6\columnwidth]{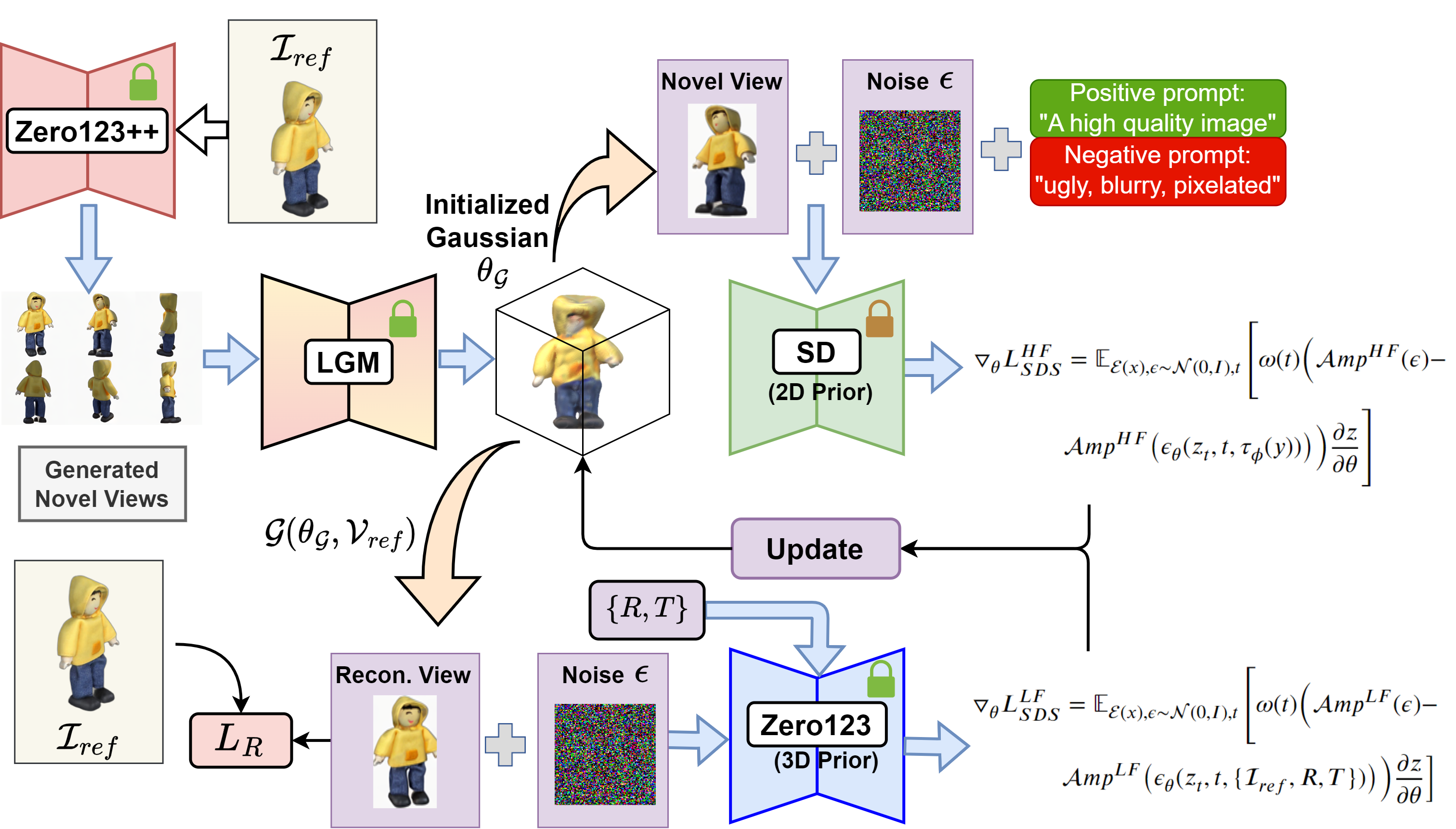}
    \caption{Overall workflow of our proposed method: First, a reference image is passed through an NVS pipeline Zero123++ \cite{shi2023zero123++} (Sec. \ref{nvs_stage}), followed by a reconstruction stage which involves a coarse Gaussian initialization using LGM \cite{tang2024lgm} and diffusion-prior based optimization  ( Sec. \ref{reconstruction}).}
    \label{fig:overall}
    \vspace{-6.4mm}
\end{figure*}

Our approach, in contrast, leverages 2D representations with the learned 3D generative priors, enabling the utilization of pre-trained 2D diffusion models with strong structural priors and 3D diffusion models with geometric prior that significantly enhance zero-shot generalization capabilities. Additionally, to address the challenge of generating multi-view consistent images, recent methods have extended 2D diffusion models to produce consistent multi-view outputs. However, these methods often struggle with low-quality geometry and blurring textures due to the inherent challenges of reconstructing 3D structures from color images alone.


Although our motivation aligns with Fourier123 \cite{yang2024fourier123}, our methodology and contributions are fundamentally distinct: (1) Instead of relying on single-image Gaussian initialization, we utilize multi-view LGM \cite{tang2024lgm} initialization, leveraging Zero123++ to generate novel views for a superior starting point; (2) Our method explicitly incorporates low-frequency components from 3D diffusion priors and high-frequency components from 2D diffusion priors—key elements absent from Fourier123’s approach; (3) We implement adaptive filtering, computing the cut-off frequency dynamically based on cumulative energy distribution in the amplitude spectrum, contrasting with Fourier123’s static amplitude-based SDS loss; (4) Sequential optimization in our framework surpasses the parallel optimization with hybrid SDS loss in Fourier123, yielding enhanced reconstruction, geometry, and texture fidelity; (5) Reference view guidance (Section \ref{reference-view-guidance}) is integrated for additional supervision, a feature not addressed in Fourier123; and (6) Our work is specifically tailored for robotics, including dedicated experiments to adapt our approach for downstream tasks, representing a significant contribution to this field.

\section{PRELIMINARIES}\label{prelim}
\subsection{3D Gaussian Splatting}
3D Gaussian Splatting (3DGS) employs a sophisticated approach to representing 3D scenes by utilizing a collection of anisotropic Gaussians. Each Gaussian is characterized by a center position $\mu \in \mathbb{R}^3 $, a scaling factor $s \in \mathbb{R}^3 $, and a rotation quaternion $q \in \mathbb{R}^4 $. Additionally, the color $c \in \mathbb{R} $ and opacity $\alpha \in \mathbb{R} $ of each Gaussian are encoded using spherical harmonic coefficients $h \in \mathbb{R}^{3 \times (k+1)^2} $ of $k$\textsuperscript{th} order. These parameters collectively define the Gaussian's properties, encapsulated in the set $\Theta = \{ \mu, s, q, c, \alpha \} $. We query 3D Gaussians as:
\begin{equation}
    G(p) = e^{-\frac{1}{2}(p-\mu)^\top \Sigma^{-1}(p-\mu)}
\end{equation}
where $p$ is the query point position. To compute the color of each pixel, a typical neural point-based rendering method is used, where the color $C[p]$ of a rendered image at pixel $p$ is given by summing the product of the color and opacity of each sampled Gaussian overlapping the pixel, taking into account the alpha composition of overlapping Gaussians as: 
\begin{equation}
    C[p] = \sum\limits_{m=1}^M c_m\alpha_m \prod_{n=1}^{m-1} (1 - \alpha_n)
\end{equation}
where $M$ Gaussians are sampled at pixel $p$; $c_m$ and $\alpha_m$ represent the color and opacity of $m$\textsuperscript{th} Gaussian.

\subsection{2D-to-3D Lifting via Score Distillation Sampling (SDS)}
DreamFusion \cite{poole2022dreamfusion} introduces score distillation sampling that leverages pre-trained text-to-image diffusion models to facilitate text-conditioned 3D generation.  Specifically, SDS optimizes the parameters $\theta$ of a differentiable 3D representation, such as a NeRF or 3DGS, by utilizing the gradient of the SDS loss $\mathcal{L}_{\text{SDS}}$ with respect to $\theta$:

\begin{equation}
\triangledown_{\theta} L_{SDS} = \mathbb{E}_{\mathcal{E}(x),\epsilon\sim\mathcal{N}(0,I),t} \left[ \omega(t)(\epsilon - \epsilon_\theta(z_t, t, \tau_\phi(y))) \frac{\partial z}{\partial \theta} \right],
\end{equation}
where $x = \mathcal{G}(\theta, \mathcal{V}) $ represents an image rendered from the 3D representation $\theta$ by the renderer $\mathcal{G} $ under a specific viewpoint $\mathcal{V}$. The weighting function $\omega(t)$ is time-dependent, and noise \( \epsilon \) is added to encoded latent $z=\mathcal{E}(x)$ at timestep \( t \), $\tau$ denotes encoder for conditioning input $y$, parameterized by $\phi$.
The core idea is to ensure that the rendered image of the learnable 3D representation aligns with the distribution of the pre-trained diffusion model. However, relying solely on 2D priors does not sufficiently capture complex 3D geometry.

Recent approaches like Magic123 \cite{qian2023magic123}, have attempted to improve generation quality by incorporating 3D priors. However, NeRF-based method suffers from two major limitations: it requires extensive time for optimization and textual inversion, and it suffers from low generalizability across views and discrepancies from the input. In contrast, our approach employs 3DGS to accelerate training, and structural supervision of appearance without requiring textual inversion, along with geometric guidance from 3D priors.


\section{PROPOSED METHOD}\label{method}
Our proposed algorithm is designed to generate an entire 3D representation from a single image in the wild. Our work necessitates a preliminary step to isolate the foreground object from the background which is done using the segmentation tool \textit{rembg} with a U2Net backbone \cite{qin2020u2}, in line with other works \cite{qian2023magic123,yang2024fourier123,shi2023zero123++}.
This is followed by a two-stage generation pipeline: (1) First, we utilize an off-the-shelf novel view synthesis (NVS) pipeline to generate multi-view images (MVIs). (2) Given the MVIs, we initialize a coarse Gaussian representation through LGM \cite{tang2024lgm}, a large reconstruction model, followed by a sequential optimization with two frequncy-based distillation losses using 2D and 3D diffusion priors. Our overall workflow is shown in  Fig. \ref{fig:overall}.

\subsection{Novel View Synthesis}\label{nvs_stage}
Our reconstruction model is designed to accept free-viewpoint images as input, allowing for the seamless integration of various multi-view generation models into our framework. This flexibility enables us to perform image-to-3D asset creation by incorporating models like MVDream \cite{shi2023mvdream}, ImageDream \cite{wang2023imagedream}, SyncDreamer \cite{liu2023syncdreamer}, and SV3D \cite{voleti2024sv3d}. We choose to use Zero123++ \cite{shi2023zero123++} for its robust multi-view consistency and tailored viewpoint distribution that effectively captures both the upper and lower regions of a 3D object, though other NVS pipelines can also be used. Leveraging its superior cross-view alignment and high-quality geometric and structural outputs, we employ Zero123++ to generate novel views at a fixed elevation across $N$ orthogonal azimuths. The novel views are represented as: 
\begin{equation}\label{nvs_generation}
    \mathcal{I}'_i = \mathcal{F}_{NVS}(\mathcal{I}_{ref}; \psi_{NVS});\forall i=\{1,2,..,N\}, 
\end{equation}
where $\mathcal{I}_{ref}$ is the input image, $\mathcal{F}_{NVS}$ is the pre-trained NVS model (Zero123++ in our case) with fixed parameters $\psi_{NVS}$.

\subsection{Reconstruction Stage}\label{reconstruction}
We employ 3DGS for its better optimization speed, and a large reconstruction model to initialize the coarse 3D representation. Specifically, inspired by recent work \cite{yi2023gaussiandreamer,chen2024text}, we initialize a coarse 3D Gaussian representation, parameterized as $\theta_{\mathcal{G}}$ using a pre-trained Large Gaussian Model \cite{tang2024lgm} with fixed parameters $\psi_{lgm}$:
\begin{equation}
    \theta_{\mathcal{G}}=\mathcal{F}_{lgm}(\mathcal{I}'_i;\psi_{lgm}); \forall i=\{1,2,..N\}
\end{equation} 
Next, we optimize $\theta_{\mathcal{G}}$ using our proposed hf-SDL.

\begin{figure}
    \centering
    \includegraphics[width=0.85\columnwidth]{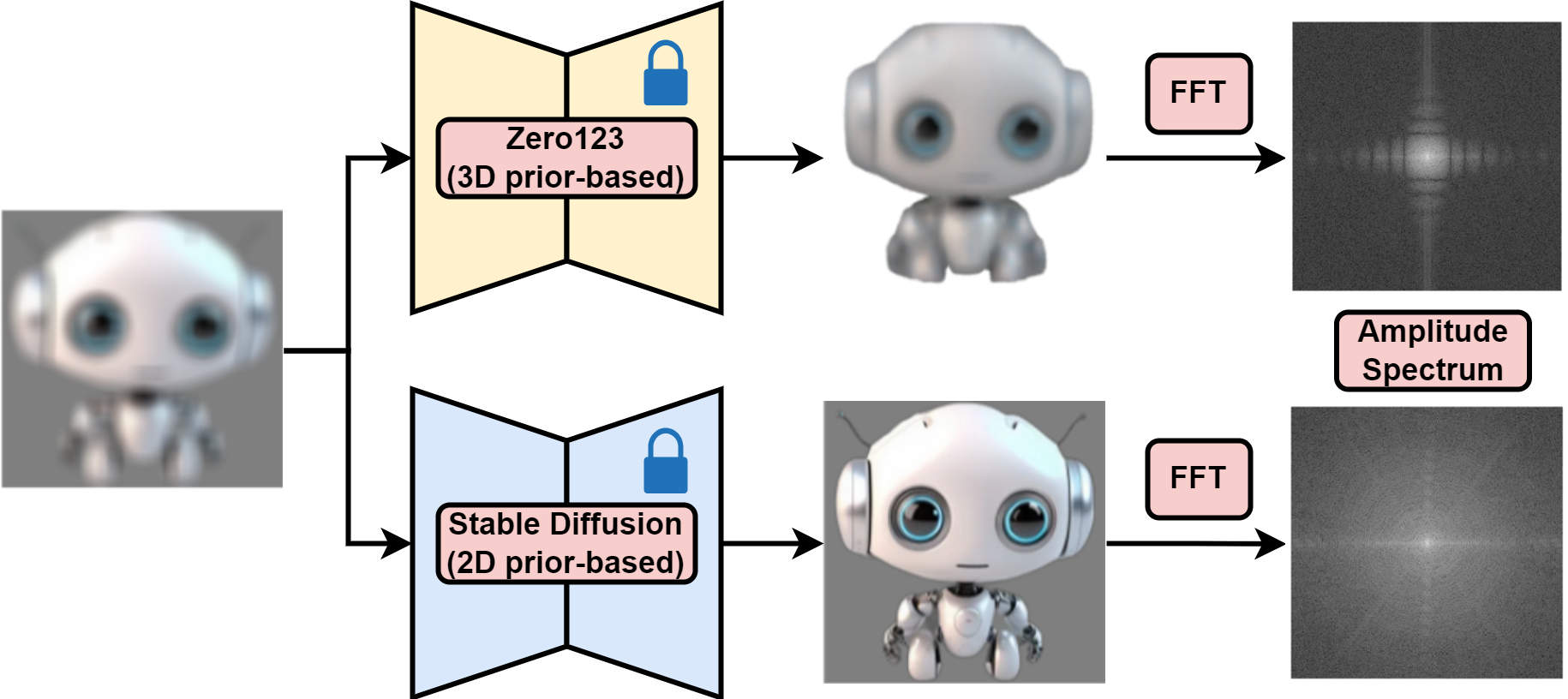}
    \caption{Amplitude analysis of 2D and 3D prior-based diffusion models: Stable Diffusion (SD) \cite{rombach2022high} and Zero123 \cite{liu2023zero}. Clearly, 3D-prior based Zero123 demonstrates blurry/smooth output, i.e., low-frequency amplitude spectrum, whereas SD demonstrates high-frequency components, i.e., sharp outputs.}
    \label{fig:2D-3D}
    \vspace{-6.4mm}
\end{figure}


\subsection{Hybrid Frequency Score Distillation Loss (hf-SDL)}\label{hf-sds-section}
The fundamental step of our proposed method is to convert an image to frequency domain from RGB. The discrete Fourier transform (DFT) is a mathematical operation that transforms a finite sequence of equally spaced samples of a function into another sequence of the same length. This new sequence consists of equally spaced samples of the discrete-time Fourier transform (DTFT), which is a complex-valued function of frequency. A multi-channel RGB image $I\in \mathbb{R}^{H\times W\times C}$ is converted to frequency domain using Fourier transformation\footnote{Channel information is omitted in the formulation for simplicity}:
\begin{equation}
    F({x,y}) =  \sum\limits_{h=0}^{H-1}\sum\limits_{w=0}^{W-1} I(x,y) e^ {-j2\pi (x\frac{h}{H} + y\frac{w}{W})}
\end{equation}
We perform this operation using an FFT algorithm \cite{frigo1998fftw} efficiently. Then the amplitude of this frequency distribution $F(x,y)$ can be computed as:
\begin{equation}
    \mathcal{A}mp(x,y) = \sqrt{\text{Real}^2(F({x,y})) + \text{Imaginary}^2(F({x,y})) }
\end{equation}
%
Our analysis of 3D-prior and 2D-diffusion models in  Fig. \ref{fig:2D-3D} reveals a stark contrast in their output quality. While 3D prior-based Zero123 \cite{liu2023zero} excels in preserving visual identity, they often generate blurred results, whereas 2D Stable Diffusion (SD) \cite{rombach2022high} produces sharper images but may hallucinate details. A spectral analysis further underscores this distinction: 3D-priors prioritize low-frequency components, ensuring identity preservation, while 2D-diffusion models emphasize high-frequency components, contributing to visual crispness. To harness the strengths of both approaches, we propose a hybrid model that leverages low-frequency components from the 3D branch for identity matching and high-frequency components from the 2D branch for structural sharpness. This synergistic combination aims to achieve a balance between fidelity and visual quality.

Specifically, we integrate the low-frequency component of 3D prior-based Zero123 model into the traditional SDS function \cite{poole2022dreamfusion} to learn from their structural and geometric knowledge, leading to identity preservation and consistent geometry in the generated asset:
\begin{equation}\label{lf_SDS}
\begin{split}
    \triangledown_{\theta} L^{LF}_{SDS} & = \mathbb{E}_{\mathcal{E}(x),\epsilon\sim\mathcal{N}(0,I),t} 
    [\omega(t)(
    \mathcal{A}mp^{LF}(\epsilon) -\\ 
    & \mathcal{A}mp^{LF}(\epsilon_\theta(z_t, t, \{\mathcal{I}_{ref},R,T\}))
    ) \frac{\partial z}{\partial \theta} ], 
\end{split}
\end{equation}
where $\mathcal{A}mp^{LF}$ represents the low-frequency component, $\{\mathcal{I}_{ref},R,T\}$ represent the reference view, rotation and translation parameters used in Zero123, respectively. 
Similarly, we adopt high-frequency component from SD branch to optimize the generated Gaussians for superior structural supervision:
\begin{equation}\label{hf_sds}
    \begin{split}
    \triangledown_{\theta} L^{HF}_{SDS} & = \mathbb{E}_{\mathcal{E}(x),\epsilon\sim\mathcal{N}(0,I),t} 
    [\omega(t)(
    \mathcal{A}mp^{HF}(\epsilon) -\\ 
    & \mathcal{A}mp^{HF}(\epsilon_\theta(z_t, t, \tau_\phi(y)))
    ) \frac{\partial z}{\partial \theta} ], 
\end{split}
\end{equation}

\begin{table}[t]
\caption{Quantitative comparison of our proposed method with the existing SOTA methods on GSO dataset. The previous best and second best performances are highlighted in \textcolor{red}{\underline{red}} and \textcolor{blue}{\textit{blue}}, whereas ours are highlighted in \hl{yellow}. }
\label{tab:GSO-comparison}
\centering
\resizebox{0.9\columnwidth}{!}{%
\begin{tabular}{@{}cccccc@{}}
\toprule
\textbf{Method} & \textbf{PSNR ↑}               & \textbf{SSIM ↑}              & \textbf{LPIPS ↓}             & \textbf{CD ↓}                & \textbf{FS ↑}                \\ \midrule
TripoSR \cite{tochilkin2024triposr}       & {\color[HTML]{FE0000} \underline{20.763}} & {\color[HTML]{FE0000} \underline{0.882}} & 0.131                        & {\color[HTML]{336EFF} \textit{0.185}} & 0.882                        \\
LGM \cite{tang2024lgm}            & 19.538                        & 0.861                        & 0.216                        & 0.345                        & 0.671                        \\
Unique3D \cite{wu2024unique3d}       & {\color[HTML]{336EFF} \textit{20.342}} & 0.879                        & {\color[HTML]{336EFF} \textit{0.128}} & {\color[HTML]{FE0000} \underline{0.184}} & {\color[HTML]{FE0000} \underline{0.886}} \\
SV3D \cite{voleti2024sv3d}           & 20.098                        & 0.861                        & 0.201                        & -                            & -                            \\
InstaMesh \cite{xu2024instantmesh}      & 19.408                        & {\color[HTML]{336EFF} \textit{0.881}} & {\color[HTML]{FE0000} \underline{0.128}} & 0.189                        & {\color[HTML]{336EFF} \textit{0.884}} \\ \midrule
\rowcolor[HTML]{FFFE65} 
\textbf{Ours}   & \textbf{22.168}               & \textbf{0.887}               & \textbf{0.121}               & \textbf{0.179}               & \textbf{0.891}               \\ \bottomrule
\end{tabular}
}
\vspace{-1.4mm}
\end{table}

\begin{table}[t]
\caption{Quantitative comparison of our proposed method with the existing SOTA methods on Omni3D dataset. The previous best and second best performances are highlighted in \textcolor{red}{\underline{red}} and \textcolor{blue}{\textit{blue}}, whereas ours are highlighted in \hl{yellow}.}
\label{tab:comparison-omni3D}
\centering
\resizebox{0.9\columnwidth}{!}{%
\begin{tabular}{@{}cccccc@{}}
\toprule
\textbf{Method} & \textbf{PSNR ↑}               & \textbf{SSIM ↑}              & \textbf{LPIPS ↓}             & \textbf{CD ↓}                & \textbf{FS ↑}                \\ \midrule
TripoSR \cite{tochilkin2024triposr}        & 19.335                        & 0.866                        & {\color[HTML]{336EFF} \textit{0.176}} & {\color[HTML]{336EFF} \textit{0.231}} & {\color[HTML]{336EFF} \textit{0.818}} \\
LGM \cite{tang2024lgm}            & 18.665                        & 0.832                        & 0.250                        & 0.356                        & 0.653                        \\
Unique3D \cite{wu2024unique3d}       & {\color[HTML]{FE0000} \underline{19.408}} & {\color[HTML]{336EFF} \textit{0.864}} & 0.181                        & 0.233                        & 0.814                        \\
SV3D \cite{voleti2024sv3d}           & 18.294                        & 0.853                        & {\color[HTML]{FE0000} \underline{0.176}} & -                            & -                            \\
InstantMesh \cite{xu2024instantmesh}    & {\color[HTML]{336EFF} \textit{19.336}} & {\color[HTML]{FE0000} \underline{0.864}} & 0.178                        & {\color[HTML]{FE0000} \underline{0.231}} & {\color[HTML]{FE0000} \underline{0.826}} \\ \midrule
\rowcolor[HTML]{F8FF00} 
\textbf{Ours}   & \textbf{20.474}               & \textbf{0.876}               & \textbf{0.168}               & \textbf{0.224}               & \textbf{0.833}               \\ \bottomrule
\end{tabular}
}
\vspace{-5.4mm}
\end{table}

\vspace{-3mm}
\subsection{Reference View Guidance}\label{reference-view-guidance}
Additionally, we propose utilizing a reconstruction loss between the reference image $\mathcal{I}_{ref}$ and rendered view at reference viewpoint $\mathcal{V}_{ref}$ as an additional supervision. Specifically, we use MSE loss by utilizing RGB image and mask:
\begin{equation}\label{reconstruction_loss}
    {L}_R = || \mathbf{A}_M \odot (\mathcal{G}(\theta_\mathcal{G}, \mathcal{V}_{ref}) - \mathcal{I}_{ref} ) ||_2^2
\end{equation}
where $\odot$ represent a Hadamard product between the foreground mask $\mathbf{A}_M$ and the rendering residual; $\theta_\mathcal{G}$ is the 3DGS parameter, used by renderer $\mathcal{G}$ at a reference view $\mathcal{V}_{ref}$.

\subsection{Overall Optimization Objective}
We optimize Gaussian representation \( \theta_\mathcal{G} \) using our proposed hf-SDS loss and $L_R$. Initially, a random 2D view is rendered from \( \theta_\mathcal{G} \), followed by noise injection and denoising via the DDIM schedule \cite{song2020denoising}, which forms the foundation of  Eq. \ref{hf_sds}. Given that the 2D diffusion prior offers high-quality yet distorted guidance, we enhance it with a 3D structural prior. In line with Zero123, we pass the rendered view along with camera intrinsics to compute the 3D SDS loss in the low-frequency domain ( Eq. \ref{lf_SDS}). Additionally, a reconstruction loss is applied to ensure reference-view fidelity (Eq. \ref{reconstruction_loss}) in combination with  Eq. \ref{lf_SDS}. 

\section{EXPERIMENTS AND RESULTS}\label{results}
\subsection{Experimental Settings}\label{experimental_setting}
We evaluate our proposed method on three widely used datasets: Google Scanned Objects (GSO) \cite{downs2022google}, OmniObject3D (Omni3D) \cite{wu2023omniobject3d}, and RealFusion15 \cite{qian2023magic123}, with $\sim1K$, $130$, and $15$ instances, respectively. Following previous works \cite{xu2024instantmesh,wu2024unique3d, long2024wonder3d}, we set $N=6$ in  Eq. \ref{nvs_generation}, normalize all generated mesh to $[-0.5, 0.5]$ bounding box to ensure alignment, and render 21 images along an orbital trajectory with uniform azimuth angles and varying elevations of $\{\ang{-30}, \ang{0}, \ang{+30}\}$ for every object. Since Omni3D also includes benchmark views sampled randomly from the top hemisphere of an object, we select 16 random views to form an additional evaluation set for Omni3D.

We optimize the 3D Gaussian ellipsoids over 500 iterations, reducing the position learning rate from \(10^{-4}\) to \(2 \times 10^{-5}\). Using SD V2 with a classifier-free guidance scale of $7.5$—sufficient for frequency-domain supervision—we avoid the quality degradation seen with higher scales (e.g., $100$). For 3D structural supervision, we employ Zero-1-to-3XL with a guidance scale of $5$, rendering at $512 \times 512$ resolution. Camera settings assume a front view (azimuth $\ang{0}$, polar $\ang{90}$) with a $1.5$-meter distance, consistent with LGM \cite{tang2024lgm}, and a $\ang{49.1}$ field of view to align with Zero123. All the experiments are performed on a single NVIDIA Tesla V100 GPU with 32GB RAM, and an entire Gaussian optimization step takes $\sim50$ seconds.  


\begin{table}[t]
\caption{Quantitative comparison of our method with the existing SOTA methods on RealFusion15 dataset. The previous best and second best performances are highlighted in \textcolor{red}{\underline{red}} and \textcolor{blue}{\textit{blue}}, whereas ours are highlighted in \hl{yellow}.}
\label{tab:comparison-realfusion}
\centering
\resizebox{0.75\columnwidth}{!}{%
\begin{tabular}{@{}cccc@{}}
\toprule
\textbf{Method} & \textbf{PSNR ↑}               & \textbf{LPIPS ↓}             & \textbf{CLIP ↑}              \\ \midrule
Make-It-3D \cite{tang2023make}     & 20.010                        & {\color[HTML]{333333} 0.119} & {\color[HTML]{336EFF} \textit{0.839}} \\
Zero123 \cite{liu2023zero}        & 25.386                        & 0.068                        & 0.759                        \\
Magic123 \cite{qian2023magic123}        & {\color[HTML]{FE0000} \underline{25.637}} & {\color[HTML]{336EFF} \textit{0.062}} & 0.747                        \\
One-2-3-45 \cite{liu2023one2345}     & 13.754                        & {\color[HTML]{000000} 0.329} & 0.679                        \\
Consistent123 \cite{weng2023consistent123}  & {\color[HTML]{336EFF} \textit{25.682}} & {\color[HTML]{FE0000} \underline{0.056}} & {\color[HTML]{FE0000} \underline{0.844}} \\ \midrule
\rowcolor[HTML]{F8FF00} 
\textbf{Ours}   & \textbf{26.244}               & \textbf{0.051}               & \textbf{0.852}               \\ \bottomrule
\end{tabular}
}
\vspace{-2.4mm}
\end{table}

\begin{figure}[t]
    \includegraphics[width=1\columnwidth]{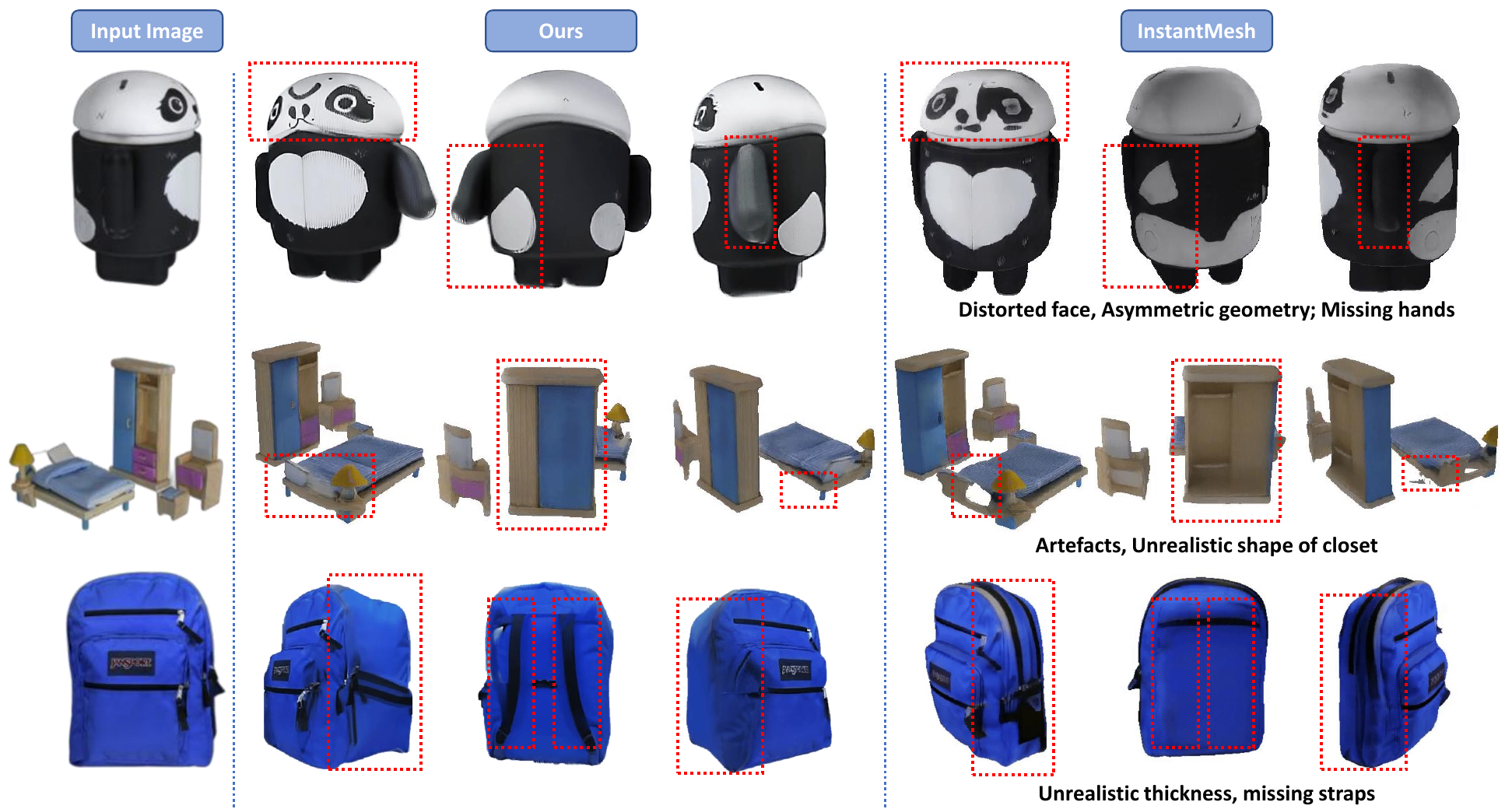}
    \caption{Qualitative comparison of our proposed method with InstantMesh \cite{xu2024instantmesh} which constantly produces unrealistic outputs with distorted face, asymmetric structure (\textit{panda}), missing realism (\textit{bag}), and artefacts (\textit{bedroom}). We encourage readers check \textcolor{red}{supplementary file} for \textcolor{red}{video results}.}
    \label{fig:InstantMesh_comparison}
    \vspace{-6.4mm}
\end{figure}

\begin{table*}[t]
\centering
\caption{Ablation experiments to assess the contribution of individual guiding components: 2D and 3D priors, along with their combinations. Our particular setting is highlighted in \hl{yellow}.}
\label{tab:ablation}
\resizebox{1.8\columnwidth}{!}{
\begin{tabular}{@{}ccccccccccccc@{}}
\cmidrule(l){1-13}
\multicolumn{3}{c}{Loss Components} & \multicolumn{2}{c}{\textbf{PSNR ↑}}                                               & \multicolumn{2}{c}{\textbf{SSIM ↑}}                                             & \multicolumn{2}{c}{\textbf{LPIPS ↓}}                                            & \multicolumn{2}{c}{\textbf{CD ↓}}                                               & \multicolumn{2}{c}{\textbf{FS ↑}}                                               \\ \cmidrule(l){1-13} 
 $L_{SDS}^{HF}$       & $L_{SDS}^{LF}$        & $L_R$       & \textbf{GSO}                            & \textbf{Omni3D}                         & \textbf{GSO}                           & \textbf{Omni3D}                        & \textbf{GSO}                           & \textbf{Omni3D}                        & \textbf{GSO}                           & \textbf{Omni3D}                        & \textbf{GSO}                           & \textbf{Omni3D}                        \\ \cmidrule(l){1-13} 
$\checkmark$       &     $\times$     &       $\times$      & 19.318                                  & \cellcolor[HTML]{FFFFFF}19.609          & 0.839                                  & \cellcolor[HTML]{FFFFFF}0.860          & 0.132                                  & \cellcolor[HTML]{FFFFFF}0.184          & 0.213                                  & \cellcolor[HTML]{FFFFFF}0.255          & 0.740                                  & \cellcolor[HTML]{FFFFFF}0.787          \\
      $\times$    & $\checkmark$       &      $\times$       & 18.771                                  & \cellcolor[HTML]{FFFFFF}17.432          & 0.787                                  & \cellcolor[HTML]{FFFFFF}0.808          & 0.139                                  & \cellcolor[HTML]{FFFFFF}0.185          & 0.192                                  & \cellcolor[HTML]{FFFFFF}0.233          & 0.851                                  & \cellcolor[HTML]{FFFFFF}0.826          \\
$\checkmark$      & $\checkmark$       &     $\times$         & 20.442                                  & \cellcolor[HTML]{FFFFFF}19.688          & 0.858                                  & \cellcolor[HTML]{FFFFFF}0.861          & 0.130                                  & \cellcolor[HTML]{FFFFFF}0.176          & 0.189                                  & \cellcolor[HTML]{FFFFFF}0.229          & 0.879                                  & \cellcolor[HTML]{FFFFFF}0.818          \\ \cmidrule(l){1-13} 
$\checkmark$       & $\checkmark$       & $\checkmark$         & \cellcolor[HTML]{F8FF00}\textbf{22.168} & \cellcolor[HTML]{F8FF00}\textbf{20.474} & \cellcolor[HTML]{F8FF00}\textbf{0.887} & \cellcolor[HTML]{F8FF00}\textbf{0.876} & \cellcolor[HTML]{F8FF00}\textbf{0.121} & \cellcolor[HTML]{F8FF00}\textbf{0.168} & \cellcolor[HTML]{F8FF00}\textbf{0.179} & \cellcolor[HTML]{F8FF00}\textbf{0.224} & \cellcolor[HTML]{F8FF00}\textbf{0.891} & \cellcolor[HTML]{F8FF00}\textbf{0.826} \\ \cmidrule(l){1-13} 
\end{tabular}
}
\vspace{-2mm}
\end{table*}

\begin{figure}[t]
    \centering
    \includegraphics[width=0.9\columnwidth]{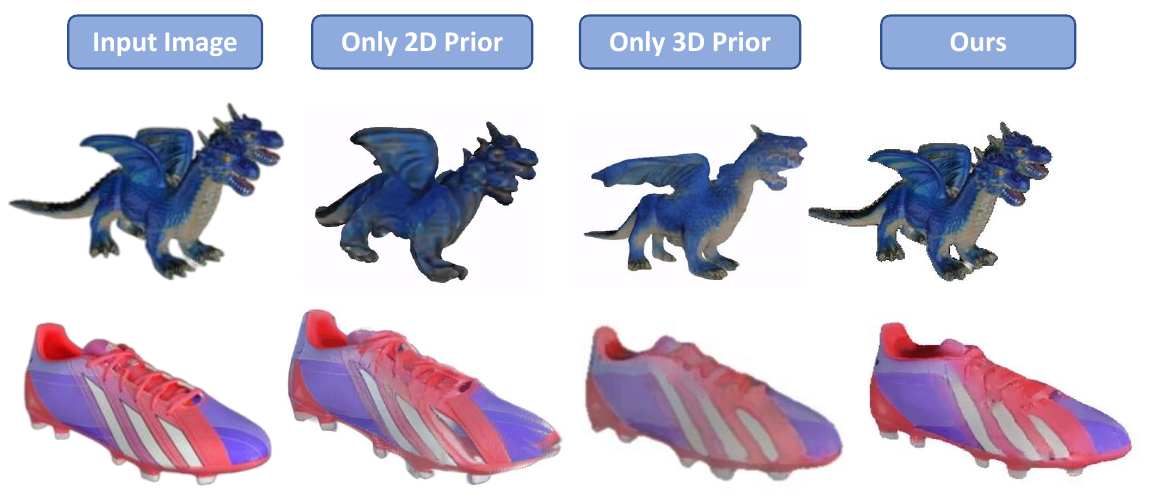}
    \caption{Qualitative ablation outcomes to assess the contribution of individual loss components. The columns represent the input image, output with only 2D prior ($L_{SDS}^{HF}$), output with only 3D prior ($L_{SDS}^{LF}$), and our proposed hybrid frequency-based SDS output, respectively.}
    \label{fig:ablation}
    \vspace{-6.4mm}
\end{figure}

\subsection{Findings and Comparison with SOTA}\label{comparison}
We assess both the 2D visual and 3D geometric quality of the generated assets. For 2D evaluation, we render novel views from the 3D mesh and compare them with the ground truth using PSNR, SSIM, and LPIPS as metrics. For 3D geometric evaluation, we align the generated meshes with the ground truth, normalize them, and then sample 16K surface points to compute Chamfer Distance (CD) and F-Score (FS) with a 0.2 threshold. Following \cite{weng2023consistent123}, we measure CLIP score for RealFusion dataset to measure 3D consistency by assessing appearance similarities across views. We compare our findings against TripoSR \cite{tochilkin2024triposr}, LGM \cite{tang2024lgm}, Unique3D \cite{wu2024unique3d}, SV3D \cite{voleti2024sv3d}, InstantMesh \cite{xu2024instantmesh}, Make-It-3D \cite{tang2023make}, Zero123 \cite{liu2023zero}, Magic123 \cite{qian2023magic123}, One-2-3-45 \cite{liu2023one2345}, Consistent123 \cite{weng2023consistent123}
in  Tab. \ref{tab:GSO-comparison},  Tab. \ref{tab:comparison-omni3D}, and  Tab. \ref{tab:comparison-realfusion} for GSO, Omni3D, and RealFusion15 datasets, respectively. Some qualtiatative visualizations are shown in  Fig. \ref{fig:our_result}.


Our methodology decisively surpasses existing state-of-the-art techniques, exhibiting marked superiority in PSNR and Chamfer Distance (CD). This underscores the robustness of our approach in harmonizing 2D structural attributes—such as nuanced color and intricate texture—with precise 3D geometry. Although our optimization-based framework operates with comparatively greater latency than inference-centric models like TripoSR, LGM, SV3D, and Zero123, it delivers a profound enhancement in both fidelity and geometric consistency, vividly illustrating the efficacy of our sophisticated two-stage coarse-to-fine paradigm. Conversely, 3D prior-based strategies, such as Make-It-3D and Unique3D, while adept at enforcing geometric consistency, are constrained in visual fidelity due to their insufficient integration of 2D prior. Visual comparison with \cite{xu2024instantmesh} in  Fig. \ref{fig:InstantMesh_comparison} justifies our superiority.

\subsection{Ablation Experiments}\label{ablation}
We perform ablation experiment to evaluate the contribution of individual loss components in  Tab. \ref{tab:ablation} and  Tab. \ref{fig:ablation}. 
\textbf{(1)} Minimizing $L_{SDS}^{HF}$, while leveraging substantial 2D information, falls short in generating realistic 3D models, as evidenced by the first column of  Fig. \ref{fig:ablation}. This limitation translates into suboptimal Chamfer Distance (CD) and F-score (FS) metrics, as detailed in  Tab. \ref{tab:ablation}.
\textbf{(2)} Conversely, $L_{SDS}^{LF}$ excels in producing geometrically accurate reconstructions but fails to capture detailed surface textures, a shortcoming clearly visible in the second column of  Fig. \ref{fig:ablation}.
\textbf{(3)} The integration of both LF and HF SDS losses results in notable advancements in both geometric fidelity and textural richness, thereby corroborating our hypothesis regarding their complementary roles.
\textbf{(4)} The addition of $L_R$, which incorporates reference view guidance, further enhances the reconstruction quality. When combined with the former losses, it achieves the most comprehensive performance, achieving superior results throughout.

\begin{figure}
    \centering
    \includegraphics[width=\columnwidth]{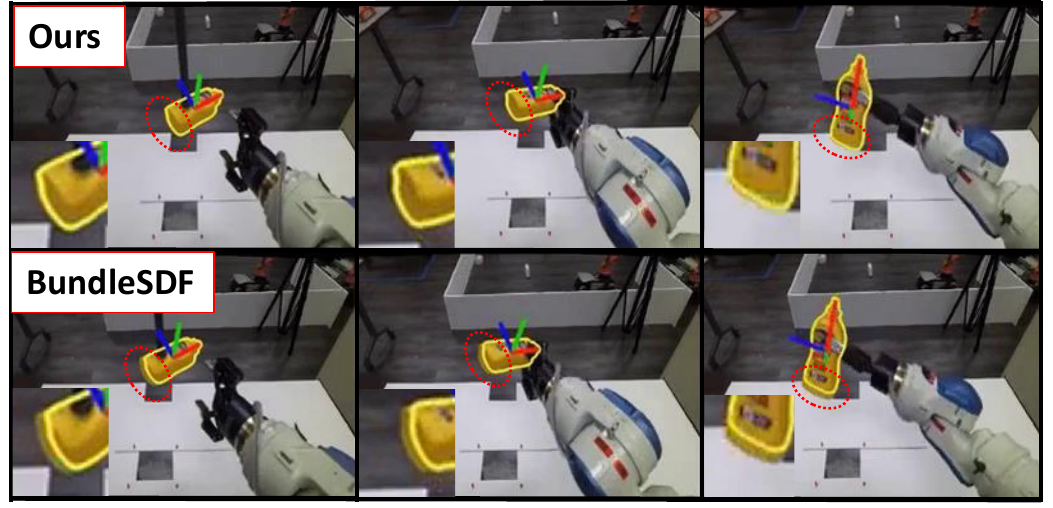}
    \caption{Comparison of pose estimation and tracking using our generated textured mesh and FoundationPose (left), and BundleSDF \cite{wen2023bundlesdf} (right). Clearly, the bounding box and tracked poses are clearly better using our generated textured mesh, as compared to BundleSDF's reconstruction pipeline (highlighted in \textcolor{red}{red}). We encourage readers to check \textcolor{red}{supplementary file} for the \textcolor{red}{video result}.}
    \label{fig:track_compare}
    \vspace{-6.4mm}
\end{figure}




\subsection{Application in Robotics}\label{application_robotics}

{Traditional model-based pose tracking achieves accurate estimation results but relies on pre-prepared textured CAD models \cite{tseng2022cla,li2023toward}, which are difficult to obtain in practice. While~\cite{wen2024foundationpose} offers an alternative by generating meshes from multi-view images, modeling diverse real-world objects remains labor-intensive, requiring manual adjustments and significant time. Although BundleSDF \cite{wen2023bundlesdf} enables simultaneous object tracking and reconstruction during testing, it requires ample of time ($\sim30$ mins.) to scan the object and optimize its 3D shape. In contrast, our pipeline generates a ready-to-use textured model from a single view in under a minute, enabling efficient downstream applications such as manipulation and tracking. We demonstrate that our complete object, generated from a single image, can be effectively used for object pose estimation and tracking when integrated with FoundationPose \cite{wen2024foundationpose}, as shown in~ Fig. \ref{fig:track_compare}.}

\section{CONCLUSION}\label{conclusion}
We propose a two-stage coarse-to-fine image-to-3D generation pipeline by effective utilization of 3D and 2D prior-based diffusion models to address the existing limitations of 3D generation. Specifically, we propose two novel frequency-based SDS loss terms, which when combined with reference-view guidance, produces state-of-the-art performance across various datasets. Moreover, we extend this pipeline for 6-DoF pose tracking of model-free in-the-wild objects, showcasing its immense potential in robotic applications.

\bibliographystyle{IEEEtran}
\bibliography{references}

\end{document}